\documentclass{article} 
\usepackage{iclr2024_conference,times}
\usepackage{graphicx}

\usepackage{amsmath,amsfonts,bm}

\def\eqref#1{equation~\ref{#1}}

\def\1{\bm{1}}

\DeclareMathAlphabet{\mathsfit}{\encodingdefault}{\sfdefault}{m}{sl}
\SetMathAlphabet{\mathsfit}{bold}{\encodingdefault}{\sfdefault}{bx}{n}

\usepackage{hyperref}
\usepackage{url}

\title{TaCo: Enhancing Cross-Lingual Transfer for Low-Resource Languages in LLMs through Translation-Assisted Chain-of-Thought Processes}

\author{Bibek Upadhayay,Vahid Behzadan, Ph.D.  \\
SAIL Lab\\
 University of New Haven\\
West Haven, CT 06516, USA \\
\texttt{\{bupadhayay,vbehzadan\}@newhaven.edu}
}

\iclrfinalcopy 
\begin{document}

\maketitle

\begin{abstract}
Creating multilingual LLMs poses a significant challenge. Pretraining or fine-tuning LLMs to adopt new languages is evidently very costly. Furthermore, there exist limitations concerning benchmark datasets and the metrics used to measure model performance in multilingual settings. This paper proposes cost-effective solutions to both aforementioned challenges. Firstly, we introduce the Multilingual Instruction-Tuning Dataset (MITS), comprised of Alpaca-52K, Dolly-15K, and Vicuna Benchmark translations into 132 languages. Secondly, we propose a new method called \emph{TaCo: Translation-Assisted Cross-Linguality}, which utilizes translations in a chain-of-thought process to instruction-tune LLMs on new languages through a curriculum-learning process. As a proof of concept, we experimented with the instruction-tuned Guanaco-33B model, performing further instruction tuning using our proposed TaCo method in three low-resource languages and one high-resource language. Our results indicate that the TaCo method impresses GPT-4 with an 82\% score for a low-resource language in the Vicuna Benchmark dataset, doubling the performance in contrast to instruction tuning alone. Furthermore, TaCo shows promise in creating multilingual LLMs, even for low-resource languages. 
We have released our datasets and model adapters\footnote{https://github.com/UNHSAILLab/TaCo}
, encouraging the research community to utilize these resources to advance work on multilingual LLMs.
\end{abstract}

\section{Introduction}
Languages are not mere tools for communication; they are rich repositories of cultural heritage, historical archives, and mediums for preserving traditions, reflecting the unique cognitive abilities of human beings. Languages encapsulate ancestral knowledge, traditions, and unique worldviews, offering insights into day-to-day lives and socio-cultural diversities. The alarming rate at which languages are disappearing, with rare languages fading into oblivion approximately every week, signals a global crisis \citep{UNESCO2022Language}. In this context, Large Language Models (LLMs) have emerged as a promising strategy for the preservation of these linguistic treasures. As the size of LLMs increases, they have shown to improve in downstream tasks not only in English but also in non-English languages \citep{goyal2021larger,openai2023gpt4, conneau2019cross} . OpenAI's GPT-4, for instance, performs across 26 languages, achieving state-of-the-art scores in the Multitask Multilingual Language Understanding (MMLU) dataset \citep{hendrycks2020measuring}.

LLMs can play a pivotal role in analyzing and revitalizing low-resource and endangered languages by teaching them vocabulary, grammar, and making use of available texts and resources. These models can facilitate the generation of additional language resources, enhance linguistic research, and create interactive applications to attract more users, thereby providing a technological buffer against language extinction. However, the availability of sufficient resources to train LLMs in non-English languages, especially those that are low-resource, remains a significant challenge.

The scarcity of rich datasets for training LLMs in non-English languages is a critical barrier. Proprietary models such as PaLM \citep{anil2023palm} and GPT-4 \citep{openai2023gpt4}, despite their capabilities in understanding and performing downstream tasks in non-English languages, are hindered by their closed model architecture and the inaccessibility of their post-processed datasets to the research community. This limitation restricts academic and independent researchers from building upon or tailoring these models for specific uses. Moreover, these proprietary models cover only a limited number of non-English languages. For instance, GPT-4's performance on languages beyond the 26 evaluated in the MMLU dataset remains unspecified, and our preliminary experiments have shown that models like GPT 3.5 often produce responses that mix similar languages, such as Sanskrit and Hindi. Similarly, models based on the newest technologies, such as Google's Bard, reveal their training limitations when tested with low-resource languages. To counteract these challenges, the open-source community has made significant strides by releasing LLMs with accessible model weights. Models such as BLOOM \citep{scao2022bloom}, POLYLM \citep{wei2023polylm}, and Glot-500 \citep{imanigooghari2023glot500} are pushing the boundaries towards truly multilingual LLMs by including non-English languages in their training datasets and employing techniques such as instruction tuning to enhance their versatility \citep{peng2023instruction}.

However, achieving multilingualism in LLMs is fraught with challenges, not the least of which is the substantial resource requirement for pretraining. Moreover, as the number of languages in a model's repertoire increases, a phenomenon known as the curse of multilinguality becomes apparent, indicating a decrease in performance for low-resource languages \citep{conneau2019unsupervised}. Aligning languages with diverse syntactic structures further complicates the matter \citep{dufter2020identifying}.

In addressing these challenges, we introduce a novel methodology named TaCo, leveraging the translation in the chain-of-thought process to foster a more inclusive multilingual model. Using established LLMs' properties—namely, emergent capabilities with increased size \citep{wei2022emergent} and the elicitation of reasoning behaviors through chain-of-thought prompting \citep{wei2022chain}—TaCo employs a curriculum learning strategy \citep{bengio2009curriculum}. It utilizes a fine-tuned Guanaco-33B model \citep{dettmers2023qlora} on the OASST1 dataset \citep{kopf2023openassistant} in conjunction with instruction tuning via Low-Rank Adaptation (LoRA)\citep{hu2021lora} for efficient fine-tuning. Distinct from other methodologies that apply LoRA adapters for multilingualism or perform instruction tuning on base models \citep{li2023bactrian,tloen_alpaca_lora}, we leverage the curriculum learning with advanced capabilities of the fine-tuned Guanaco-33B model. This approach streamlines the process of teaching the model to translate and generate responses in respective languages, minimizing the need for intensive model training from scratch and thereby saving on overall training costs. Through these efforts, we aim to contribute to preserving the linguistic diversity that enriches the fabric of human culture and cognition.

We present our overall contribution as follows:
\begin{enumerate}

    \item We present the Multilingual Instruction-Tuning Dataset (MITDS), which consists of the translated Alpaca-52K \citep{peng2023instruction} and Dolly-15K \citep{DatabricksBlog2023DollyV2}. These datasets, translated using Google Cloud Translation, cover 132 languages. 
    \item We propose a new method called TaCo for instruction-tuning LLMs to learn new languages. 
    \item We also present the Multilingual Vicuna Benchmark, which is a dataset comprised of  translation of the Vicuna Benchmark \citep{chiang2023vicuna} in 132 languages and made them publicly available.
    \item We evaluate four TaCo models on the Vicuna Benchmark, for 3 low-resource languages (Nepali, Sanskrit, and Maithili), as well as a high-resource language: Persian. 
    \item We release the adapters for the aforementioned four language models for public use. 
   
\end{enumerate}

\section{Related Work}
Training LLMs for multilingual purposes involves two major approaches: pretraining on multilingual data and fine-tuning in a new language. \cite{kenton2019bert} introduced the concept of a multilingual-BERT pretrained model , which laid the foundation for subsequent advancements. \cite{conneau2019unsupervised} discussed the trade-offs involved in adding languages to improve performance on low-resource languages, called \emph{curse of multilinguality} . They noted that past a certain point, performance begins to decline. Their work with the XLM-R model involved training on extensive data to bolster cross-lingual representation. Improving upon these efforts, \cite{chi-etal-2022-xlm} introduced discriminative pretraining tasks in the creation of XLM-E, which enhances cross-lingual transferability while reducing computational cost. \cite{dufter2020identifying}'s approach utilized smaller BERT models, incorporating synthetic data alongside natural data, and adjusted the masking strategy to create efficient multilingual BERT models .

Considering the impact of model size, Goyal et al. trained XLM-RXL and XLM-RXXL models, demonstrating improved cross-lingual understanding in larger models \cite{goyal2021larger}. Their findings highlight the capabilities of large-scale models in zero-shot learning and underscore the enhanced performance achievable through fine-tuning. \cite{scao2022bloom} introduced BLOOM, a large 176B-parameter open-source model, trained on a diverse corpus to support multiple languages and tasks, showing remarkable task generalization. Following this, \cite{muennighoff2022crosslingual} furthered this initiative by fine-tuning BLOOM and mT5 models on multitask prompts, resulting in the BLOOMZ and mT0 variants that demonstrated improved performance across languages and tasks. \cite{li2023bactrian} embarked on training Bactrian-X models using the LoRA approach, aiming at bridging language representation in a vast array of languages. Similarly, \cite{wei2023polylm} released POLYLM, a model trained on a large multilingual dataset, employing a curriculum strategy to foster both commonsense generalization and multilingualism. \cite{imanigooghari2023glot500} targeted predominantly low-resource languages with the Glot500-m model, creating a tailored pretraining environment to evaluate multilingual models' performance effectively . Lastly, \cite{kudugunta2023madlad} explored the creation of a vast multilingual machine translation dataset, underpinning training models with up to 10.7B parameters, focusing on minimizing errors through a mix of original and back-translated data.
\section{Experiment}
\label{sec:dataset}

\textbf{Dataset: }We initially translated two datasets, Alpaca-52K-GPT4 and Dolly-15K, into 132 languages using the Google Cloud Translation service. Due to computational resource constraints, we focused our experiment on three low-resource languages—Sanskrit, Nepali, and Maithili—and one high-resource language, Persian. Aware of the risk of \emph{translationese}, a common issue in machine translations where the output deviates from native linguistic norms through literal translations, foreign syntax, and odd phrases \citep{Borah_Translationese_2021, koppel2011translationese}, we conducted a manual evaluation for translation quality. We selected a sample of 1000 sentences from each language and assessed them using the BLEU score, calculated with ScarceBLEU \citep{post2018call}, ChrF \citep{popovic2017chrf}, and the Translation Error Rate (TER) \citep{snover2006study}, by translating from English to the target language and back (en-xx-xx-en).

\begin{figure}[h]
\begin{center}
\includegraphics[width=5.5in]{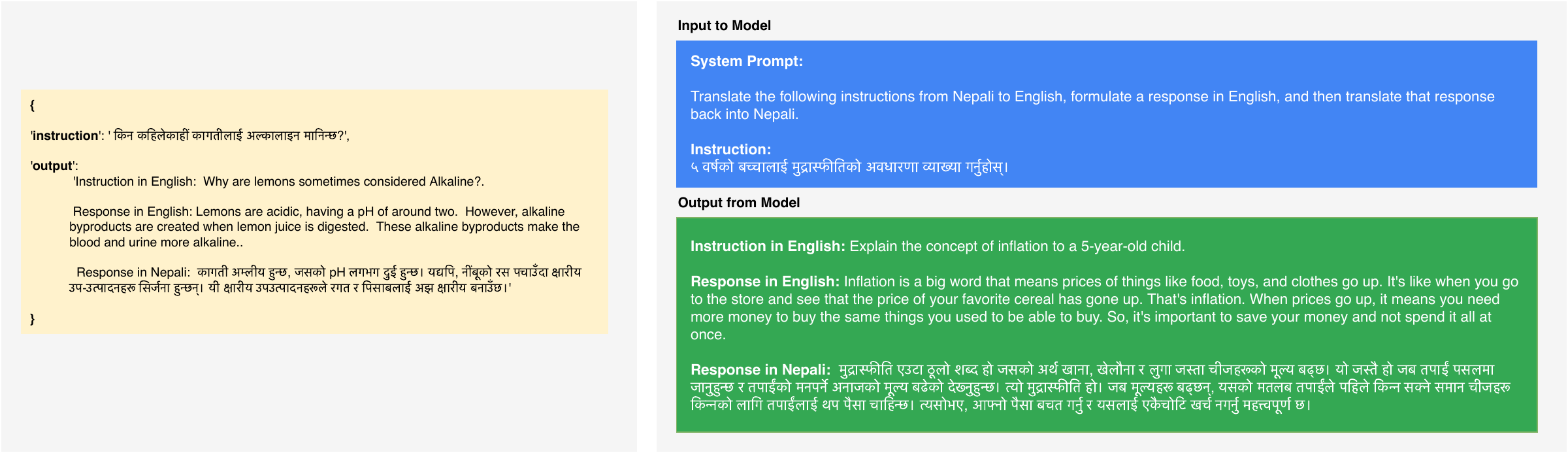} 
\end{center}
\caption{Left: An example of a data point in the TaCo Dataset. Right: Instruction and the response generated using TaCo method.}
\label{fig:prompt_point}
\end{figure}

In our multilingual fine-tuning process, we employed a chain-of-thought approach to transform the translated instruction dataset. This method involves breaking down the complex problem of cross-linguality into manageable reasoning steps, akin to a chain of thought. By combining the English dataset with another language, we crafted question-answer pairs called as TaCo datasets. Here, instructions are given in a non-English language, while outputs are provided in both English and the original language. The process unfolds by first translating the instruction into English, formulating the response in English, and then translating this back into the non-English language. We avoid translating the \emph{input} from the target language (xx) into English in our \emph{output}. This method fosters the model's comprehension and utilization of the original language nuances. By combining the Alpaca-52K and Dolly-15K datasets, we have curated a comprehensive collection of 67K question-answer prompts. The example of a TaCo data point is given in Fig. \ref{fig:prompt_point} (Left).

\textbf{Training: }Our training follows a curriculum learning strategy where we train the \emph{pre-finetuned} model which has already produced state-of-the-art results in generating responses in English. Accordingly, we selected the Gaunaco-33B fine-tuned model and further instruction-tuned it with our TaCo method. We transformed the dataset to align with our TaCo method as mentioned earlier. The hyperparameters for the model's training are as follows: a learning rate of 3e-4 and a cutoff length set to 2000. Our training targeted four LoRA modules: q\_proj, k\_proj, v\_proj, and o\_proj, with the LoRA parameters r, alpha, and dropout set at 32, 64, and 0.05, respectively. 


\textbf{Evaluation: }In our model evaluation we used Vicuna Benchmark \cite{chiang2023vicuna}, however, the benchmark is available only in English. We translated the benchmark dataset into 132 languages using Google Cloud Translation and have made them publicly available. To evaluate our TaCo models, we followed the single answer evaluation method in which we asked the GPT-4 model to judge the answers generated by our models. We generated answers for the questions in the following four languages: Nepali, Persian, Sanskrit, and Maithili. An example of the output generated  from the model is depicted in the Fig. \ref{fig:prompt_point} (Right).

As we can see, the input to the model is in the native language, with the system prompt provided in English. This is because the underlying model being used was fine-tuned in the English language and is therefore capable of understanding the system prompt and acting accordingly. In the model's response, we can observe that the instruction in Nepali is initially translated to English, and then a response is generated in English followed by a response in Nepali. One of the main challenges faced in this approach is the token limit while generating responses. Open-ended question responses are generally longer, and when implemented using the TaCo method, need to be first generated in English and then translated into the respective language, often exceeding the maximum token limit in the LLaMA model.

This typically results in longer English response generation and incomplete response generation in the native language due to model token limitations. We resolved this by appending a simple instruction to the prompt, specifying that the model response should not exceed six sentences. After exhaustive trials, we found that six sentences is the sweet spot. Moreover, we also modified the system prompt in the evaluation benchmark by adding an extra line: 'Do not allow the length of the responses to influence your evaluation', which will make the GPT-4 not judge based on the length of the answer. 

\section{Results}

Our Vicuna Benchmark results are summarized in Table \ref{table:evaluation_results}, detailing performance by category and language. The average scores were 88\% for Nepali, 80\% for Sanskrit, 82\% for Maithili, and 84\% for Persian across nine categories. Except for math, where the Sanskrit model scored the lowest, all TaCo models achieved scores above 70\%. We excluded coding from the evaluation since programming languages are usually in English, allowing for a more accurate assessment of model performance.  

To evaluate the effectiveness of the TaCo method, we conducted a comparative analysis by instruction-tuning the Guanaco-33B model on two languages of varying resource availability—Nepali, a low-resource language, and Persian, a high-resource language. The instruction-tuning was performed utilizing translations of the Alpaca-52K and Dolly-15K datasets into the respective languages, maintaining consistency in hyperparameters across both models. The results, as illustrated in Table \ref{table:evaluation_results}, demonstrate that our TaCo method markedly surpasses traditional approaches, nearly doubling the average performance score.

\begin{table}[t]
\caption{Comparison of scores (in \%) of TaCo models vs. Instruction-Tuned (IT) models.}
\label{table:evaluation_results}
\begin{center}
\begin{tabular}{lllllll}
\multicolumn{1}{c}{\bf Category}  &\multicolumn{1}{c}{\bf Nepali} &\multicolumn{1}{c}{\bf Sanskrit} &\multicolumn{1}{c}{\bf Persian} &\multicolumn{1}{c}{\bf Maithili} &\multicolumn{1}{c}{\bf Nepali-IT} &\multicolumn{1}{c}{\bf Persian-IT}
\\ \hline \\
Common Sense & 85.5 & 83.0 & 86.0 & 78.0 & 40.0 & 63.0 \\
Counterfactual & 89.0 & 83.0 & 84.6 & 83.0 & 18.0 & 46.0\\
Fermi & 77.0 & 77.0 & 82.0 & 75.0  & 31.0& 30.0 \\
Generic & 90.0 & 91.3 & 90.0 & 93.0 & 63.0 & 44.0 \\
Knowledge & 95.5 & 94.2 & 92.5 & 88.0 & 47.0 & 52.5 \\
Math & 96.7 & 33.3 & 63.3 & 70.0 & 43.3  & 56.7 \\
Roleplay & 83.5 & 89.0 & 86.0 & 78.0 & 36.0  & 38.5\\
Writing & 88.0 & 89.5 & 86.5 & 93.0 & 36.0 &  41.0 \\
\bf Overall Average & \bf 88.1 & \bf 80.0 & \bf 83.9 & \bf 82.2 & 39.3 & 46.5  \\
\end{tabular}
\end{center}
\end{table}

\section{Discussion }

The LLMs performance increases with their sizes and is capable of solving the multilingual grade school math problems using chain-of-thought process, which can also be applied to other common-sense problems \citep{shi2022language, wei2022chain}. Similarly, we initially experimented with few-shot chain-of-thought examples in our base Guanaco-33B model, but the model was unable to generate non-English response. One of the probable reason could be the lack of exposure to the language during pretraining and finetuning. Nonetheless, further LoRA instruction tuning revealed the model's capability to generate responses in that particular language. But instruction-tuned model, capable of generating responses in non-English languages, encountered several issues such as broken grammar, lack of information (like facts, history, and significant events), hallucinations, and repetitive sentences during generation. However, most of these issues were resolved when we implemented our proposed TaCo method. This improvement is evident in our average results, which display an accuracy of 80\% or above for each language and an average accuracy of 80\% in the common sense category across all four languages.

The curriculum learning strategy emphasizes starting with basic principles and advancing to more complex topics. We adopt this approach in teaching models to respond in English before tackling low-resource languages, using the state-of-the-art Guanaco-33B model. This mirrors human second-language learning—initially analyzing and responding in one's native language before switching to English. Our TaCo method reflects this by first translating to English, generating the response, and then translating back to the non-English language, adhering to a chain-of-thought process that simplifies complex problems through sequential tasks, with translation at both the beginning and end. This approach capitalizes on the pre-training knowledge of the English language to generate responses in other languages, utilizing its understanding to provide more accurate and factual information while minimizing errors. Consequently, it achieves an average accuracy of 92\% in the knowledge category across four languages.

Despite its proficiency in handling non-English questions with an impressive 83\% success rate by GPT-4 standards, the model faces significant limitations. One major issue is the token limit, as responses in both languages often exceed the allowed number of tokens, restricting the length of the responses. Furthermore, the model's creativity in the native language diminishes, limiting its ability to craft poetry or rhymes in the target language, as it operates primarily from an English perspective. Another challenge is the increased token count and time required for generating responses in non-English languages, which we observed to be particularly true for the four languages we tested. This not only increases costs but also the time needed for response generation.

\section{Conclusion and Future work}

In this study, we introduced TaCo, a novel approach for crafting multilingual models through translation in the chain-of-thought process, impressively impacting GPT-4 with an 82\% accuracy on the Vicuna Benchmark dataset. TaCo effectively generates multilingual models at a reasonable cost, with model weights and datasets now publicly available. Despite its advancements, the study recognizes limitations, including a lack of robustness, absence of toxicity tests, and challenges posed by token limits. The large size of our released models indicates that future efforts will aim for efficiency in smaller models. In conclusion,
the emergent behavior of LLMs, amplified by translation in the chain-of-thought process, can enable multilingualism within these models.
\bibliography{iclr2024_conference}
\bibliographystyle{iclr2024_conference}

\appendix
\section{Appendix}

\subsection{Translation Evaluation Metrics }
We present the translation evaluation metric scores for three low-resource languages—Sanskrit, Nepali, and Maithili—alongside Persian, a high-resource language, in Table \ref{table:translation_evaluation_metrics}.

\begin{table}[t]
\caption{Translation Evaluation Metrics for Low Resources Languages}
\label{table:translation_evaluation_metrics}
\begin{center}
\begin{tabular}{llll}
\multicolumn{1}{c}{\bf Language}  &\multicolumn{1}{c}{\bf BLEU} &\multicolumn{1}{c}{\bf CHRF++} &\multicolumn{1}{c}{\bf TER}
\\ \hline \\
Sanskrit & 65.23 & 84.62 & 19.43 \\
Nepali & 69.68 & 87.37 & 15.02 \\
Persian & 62.42 & 80.72 & 20.61 \\
Maithili & 63.65 & 84.88 & 19.58 \\
\end{tabular}
\end{center}
\end{table}

\subsection{Evaluation Loss for TaCo Models}

\begin{figure} 
\begin{center}
\includegraphics[width=1\linewidth]{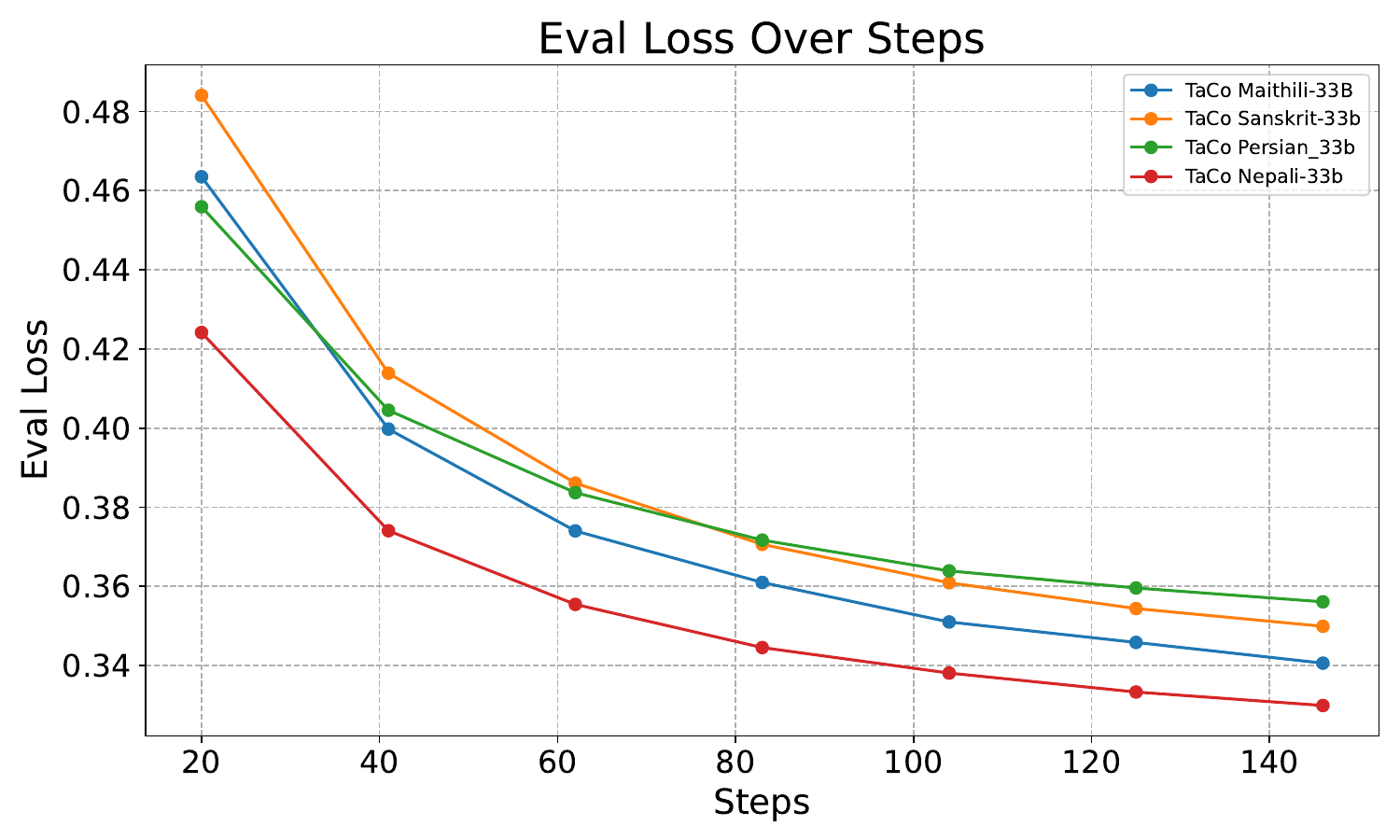} 
\end{center}
\caption{Evaluation Loss over steps for four different TaCo models.}
\label{fig:evaluation_loss}
\end{figure}

In Fig. \ref{fig:evaluation_loss}, we plot the evaluation loss for four models. As can be seen, there is a steady decline in the loss throughout the steps in all four models.

\subsection{Comparison of TaCo Models VS Instruction-Tuned Model}

\begin{figure}
\begin{center}
\includegraphics[width=1\linewidth]{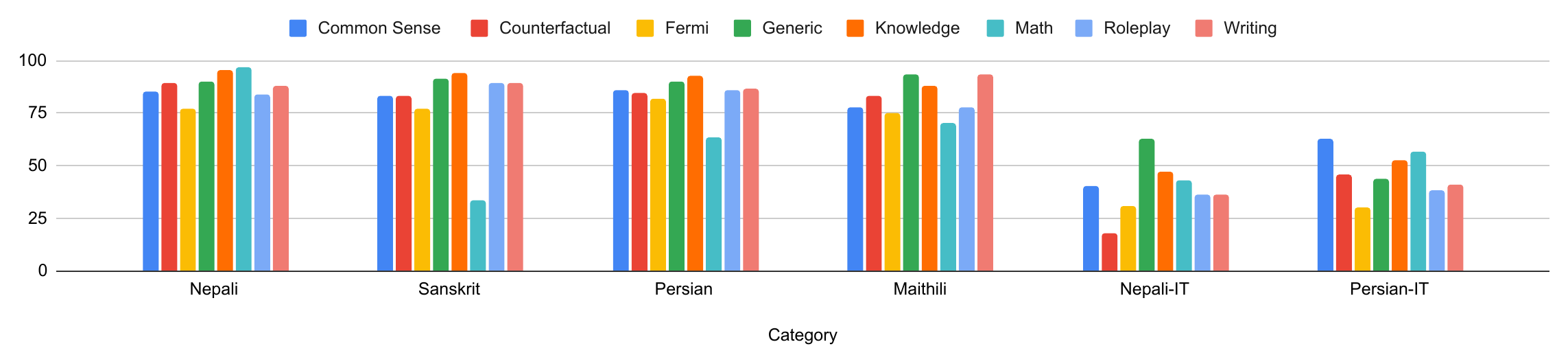} 
\end{center}
\caption{: Comparison of Vicuna Benchmark scores by categories for different TaCo models and the
Instruction Tuned (IT) model.}
\label{fig:each_model_performance}
\end{figure}

\begin{figure} 
\begin{center}
\includegraphics[width=1\columnwidth]{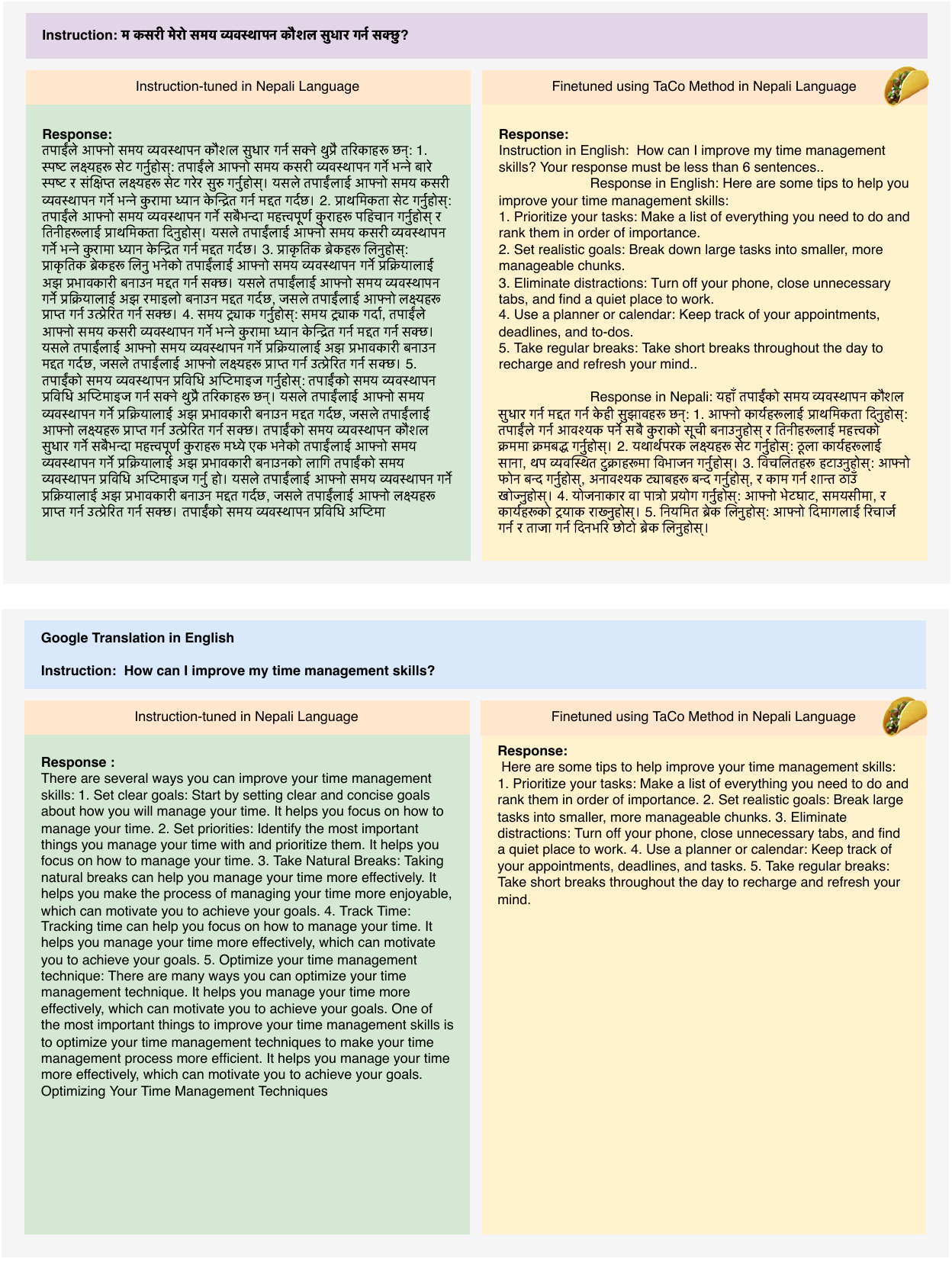} 
\end{center}
\caption{Example response from models on 'How can I improve my time management skills?' }
\label{fig:q1}
\end{figure}

\begin{figure} 
\begin{center}
\includegraphics[width=1\columnwidth]{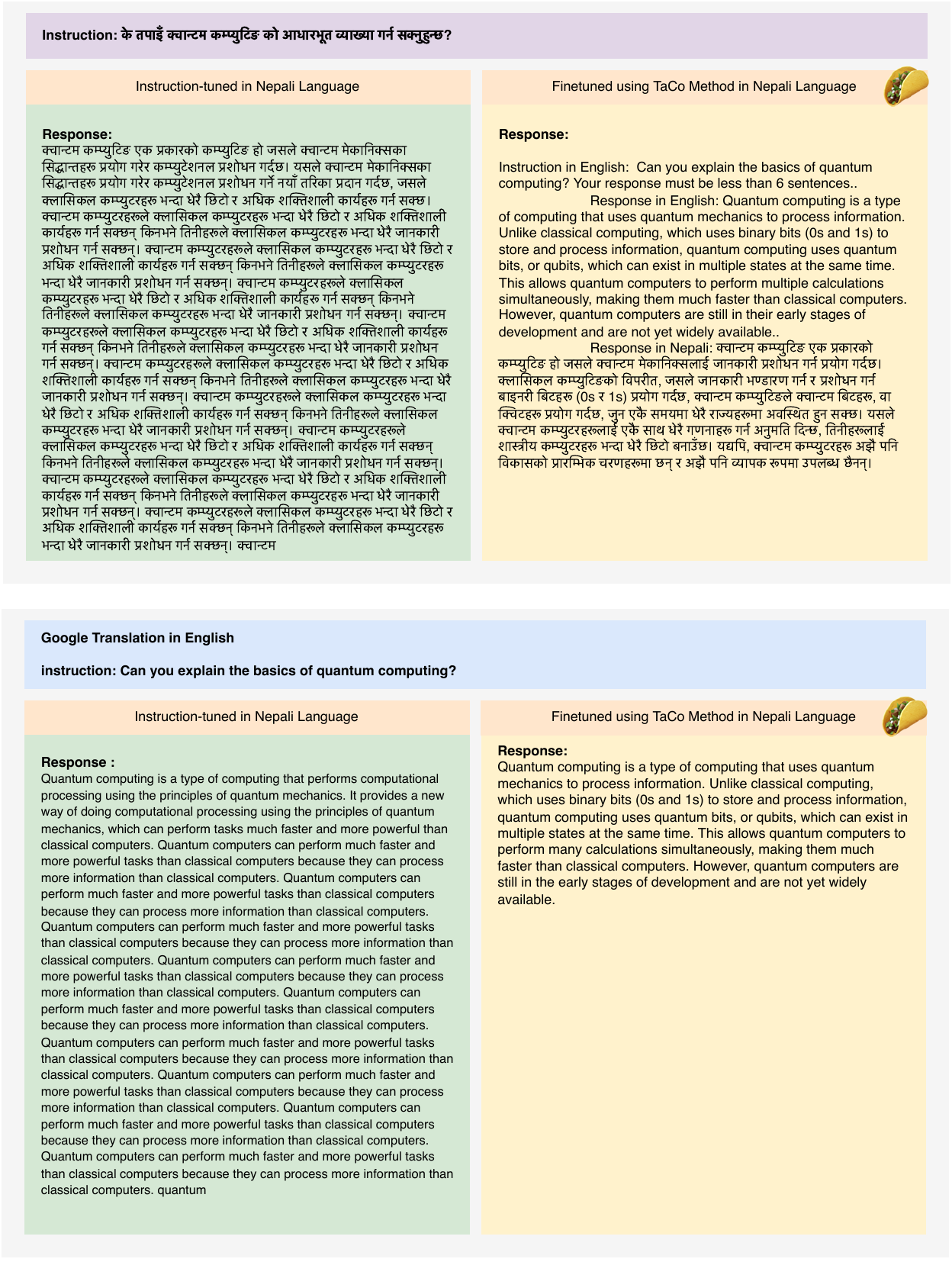} 
\end{center}
\caption{Example responses from models to the question 'Can you explain the basics of quantum computing?'  }
\label{fig:q5}
\end{figure}

\begin{figure} 
\begin{center}
\includegraphics[width=1\columnwidth]{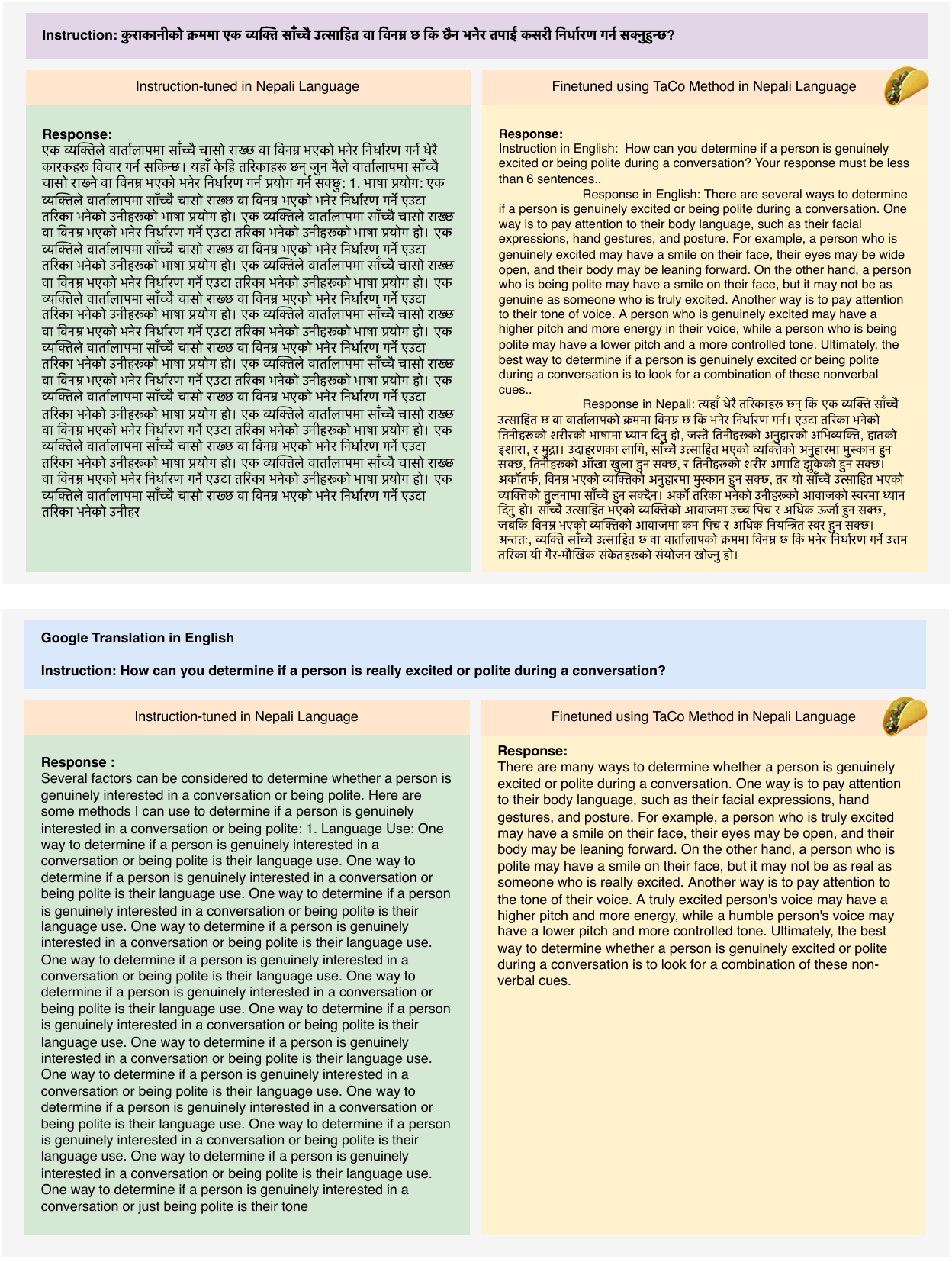} 
\end{center}
\caption{Example responses from models to the question 'How can you determine if a person is really excited or polite during a conversation?'  }
\label{fig:q34}

\end{figure}

In Fig.\ref{fig:each_model_performance}, we plot the scores of categories from each model. We observe that the instruction-tuned-only models perform inferiorly compared to the TaCo Models. In Fig. \ref{fig:q1}, Fig. \ref{fig:q5}, and Fig. \ref{fig:q34}, we illustrate responses from the instruction-tuned model in Nepali alongside those from the TaCo Nepali model. We also provide a Google translation in English for each response. We appended the instruction \emph{'Your response must be less than 6 sentences'} to ensure that the model's response does not exceed the token limit. In the responses from the instruction-tuned-only model, it can be observed that the model begins to provide an answer but starts repeating the same answer after a few sentences. In contrast, the responses from the TaCo models are concise and to the point. 

\end{document}